\documentclass[sn-apa]{sn-jnl}% APA Reference Style 
\usepackage{graphicx}%
\usepackage{multirow}%
\usepackage{amsmath,amssymb,amsfonts}%
\usepackage{amsthm}%
\usepackage{mathrsfs}%
\usepackage[title]{appendix}%
\usepackage{xcolor}%
\usepackage{textcomp}%
\usepackage{manyfoot}%
\usepackage{booktabs}%
\usepackage{algorithm}%
\usepackage{algorithmicx}%
\usepackage{algpseudocode}%
\usepackage{listings}%

\begin{document}

\title[Automating Patent Response: PARIS \& LE-PARIS Systems]{From PARIS to LE-PARIS: Toward Patent Response Automation with Recommender Systems and Collaborative Large Language Models}


\author[1,2]{\fnm{Jung-Mei} \sur{Chu}}\email{D09944017@csie.ntu.edu.tw}

\author*[1,2]{\fnm{Hao-Cheng} \sur{Lo}}\email{austenpsy@gmail.com}

\author[2]{\fnm{Jieh} \sur{Hsiang}}\email{hsiang@csie.ntu.edu.tw}

\author[1]{\fnm{Chun-Chieh} \sur{Cho}}\email{Jeff@jcipgroup.com}

\affil[1]{\orgname{JCIPRNET}, \orgaddress{\city{Irvine}, \state{CA}, \country{USA}}}

\affil[2]{\orgname{National Taiwan University}, \orgaddress{\city{Taipei}, \country{Taiwan}}}


\abstract{In patent prosecution, timely and effective responses to Office Actions (OAs) are crucial for securing patents. However, past automation and artificial intelligence research have largely overlooked this aspect. To bridge this gap, our study introduces the Patent Office Action Response Intelligence System (PARIS) and its advanced version, the Large Language Model (LLM) Enhanced PARIS (LE-PARIS). These systems are designed to enhance the efficiency of patent attorneys in handling OA responses through collaboration with AI. The systems' key features include the construction of an OA Topics Database, development of Response Templates, and implementation of Recommender Systems and LLM-based Response Generation. To validate the effectiveness of the systems, we have employed a multi-paradigm analysis using the USPTO Office Action database and longitudinal data based on attorney interactions with our systems over six years. Through five studies, we have examined the constructiveness of OA topics (studies 1 and 2) using topic modeling and our proposed Delphi process, the efficacy of our proposed hybrid LLM-based recommender system tailored for OA responses (study 3), the quality of generated responses (study 4), and the systems' practical value in real-world scenarios through user studies (study 5). The results indicate that both PARIS and LE-PARIS significantly achieve key metrics and have a positive impact on attorney performance.}




\maketitle

\section{Introduction}\label{sec1}

To address the increasing volume and complexity of inventions filed with the United States Patent and Trademark Office (USPTO) over the last two decades \citep{USPTOFY2023}, automation, language model (LM), and artificial intelligence (AI) technologies have been extensively applied in patent drafting and prosecution, including tasks such as patent classification \citep{lee2020clas} and claim generation \citep{lee2020claim}, as well as early stages of patent examination like prior art search \citep{helmers2019automating} and patentability assessment \citep{lo2021pre}.

Despite these efforts, automation has not thoroughly benefited one critical aspect of the patent prosecution life cycle: the Office Action (OA) and response to OA process. This process requires detailed communication and extensive exchanges of technical and legal knowledge among examiners, inventors, and patent agents (attorneys). Such ongoing negotiation and effective communication are instrumental in resolving technical and legal issues, thereby enhancing the likelihood of patent approval \citep{tu2021fast}.

The response process in patent prosecution is inherently time-sensitive. In patent law, the importance of timeliness is underscored by statutory response deadlines. Failure to respond promptly can render an application invalid, adversely affecting the applicant's patent rights \citep{tu2021fast}. For inventors, timeliness directly influences the market value of their innovations. Expedient responses can accelerate the patent approval process, allowing inventors to enter the market more swiftly and sustain market competitiveness \citep{gaudry2012lone}. Patent agents, snowed under multiple cases, must adhere to time constraints to facilitate seamless case progression and avoid workload overflow, impacting the quality of client service and case management capabilities \citep{de2023patent}. Lastly, examiners benefit from the applicants' timely responses in maintaining the review process's pace and efficiency \citep{lu2017uspto}.

Given the urgency to streamline and assist inventors or patent attorneys in the response to OA, we introduce the \textit{Pattern Office Action Response Intelligence System (PARIS)}. Developed in 2018 and continuously updated, PARIS semi-automatically provides a case-oriented and user-need-based \textit{response template} recommender system, leveraging LM techniques. Its core objective is to offer attorneys precise strategies to handle various OA scenarios, thereby accelerating the patent examination process.

With the advent and advancement of Large Language Models (LLMs), we in 2023 further propose the \textit{LLMs Enhanced PARIS (LE-PARIS)}. LE-PARIS, building upon the existing PARIS framework, incorporates the powerful capabilities of LLMs to enhance the automation of response generation and the accuracy of language processing. The system primarily introduces an LLM-based recommender system, integrating response templates with external relevant resources and user-defined remark of any style or language to directly generate responses that meet the linguistic standards and requirements of the USPTO.

To evaluate the efficacy of the systems aforementioned, our research raises several fundamental questions. For PARIS, we investigate whether the response templates are well-constructed (Studies \hyperref[sec:study1]{1} and \hyperref[sec:study2]{2}). We also examine the performance of the recommendater system in terms of several accuracy metrics (Study \hyperref[sec:study3]{3}). For LE-PARIS, we explore whether the integration of LLMs enhances the effectiveness of the recommendater system (Study \hyperref[sec:study3]{3}) and the quality of the generated responses (Study \hyperref[sec:study4]{4}). Finally, we assess the practical utility of both systems for users (Study \hyperref[sec:study3]{5}). Hence, our contributions to the field are three-fold:
\begin{enumerate}
  \item We pioneered the development of PARIS and LE-PARIS to facilitate the crucial, yet underappreciated, process of responding to Office Actions.
  \item Both systems synergize our proposed response template recommender systems with advanced LM and/or LLM technologies, adopting a collaborative approach. This methodology addresses complex legal and technical content in patents, moving beyond mere automation with inaccuracies.
  \item We utilize a multi-paradigm approach for scientific validation of the systems' effectiveness, not only proving their efficacy through scientific metrics but also demonstrating their practical value.
\end{enumerate}

\section{Preliminaries and Background}\label{sec2}

For automating the OA Response process, we initially focus on the use of \textit{response templates}. Despite the complex legal and technical terms involved, patterns can still be discerned under various rejection statutes, making templates an effective framework or strategic guideline. Previous attempts have highlighted the significant benefits of templates in organizing and completing patent-related and legal documents \citep{markovic2023legal}. For instance, provisional application templates provide a structured outline guiding applicants to correctly complete provisional patent applications, even without attorneys' assistance \citep{Erickson2012Provisional}. Moreover, templates have been widely used in drafting various legal documents, such as contracts or confidentiality agreements. Such templates can be customized to individual needs, ensuring consistency and professionalism in documents while saving drafting time \citep{Wernikoff2020Templates}. Therefore, we construct high-quality templates for responding to OAs. These templates are able to guide applicants or agents to cover all necessary information and effectively organize response structures. The response templates feature fill-in-the-blank to facilitate customization, ensuring that responses not only comply with patent legal requirements but also accurately reflect the unique characteristics of each case.

While templates enhance the efficiency of patent attorneys in drafting responses, the vast array of OA topics and their corresponding templates makes finding an appropriate template a potential bottleneck in automation. To address this, we consider the integration of recommender systems. These systems ease the way of decision-making and provides personalized suggestions based on user behaviors and the OA content, primarily hybridising collaborative filtering (CF) and content-based filtering (CB) approaches. CF recommends items (i.e. response template) based on the preferences of similar users, while CB makes recommendations based on item features \citep{su2009survey, wu2021survey}. In fact, recommender systems have been applied in patent and legal studies \citep{domingues2023large}. For example, \cite{chen2023interpretable} has primarily introduced an interpretable patent recommendation method, capable of recommending patents based on their features. Thus, we utilize recommender systems to assist patent attorneys in finding suitable response templates and strategies to address various issues in OAs.

Given the fundamentally text-based nature of patent prosecution, the advent of Natural Language Processing (NLP), LM, and LLM has opened new avenues in this domain \citep{krestel2021survey, lee2023evaluating}. These advanced technologies exhibit remarkable capabilities in contextual reasoning, precise text representation, and efficient text generation \citep{zhao2023survey}. These strengths are able to provide substantial support in text processing, recommendation system development, and response generation within our framework. The direct application of these technologies in crafting OA responses, however, remains an understudied area.

Since 2022, LLMs such as GPT-3, GPT-4, and LLaMA have shown distinctive in-context learning abilities and ability eliciting \citep{zhao2023survey}. While these generative models have been employed in drafting patent claims and pretrained on patent texts \citep{lee2023evaluating}, producing complete and high-quality patent content is still challenging. This challenge largely hinges on precise user input and effective prompt engineering (i.e. \textit{garbage in, garbage out}). Therefore, we investigate the potential of a human-AI collaborative approach \citep{wang2020human}, focusing on how patent attorneys can synergistically utilize the technical prowess of LLMs to elevate the quality and efficiency of OA responses.

\section{System Overview}\label{sec:overview}

\begin{figure}[ht]
\centering
        \hspace*{-80pt}
	  \includegraphics[width=1.5\textwidth]{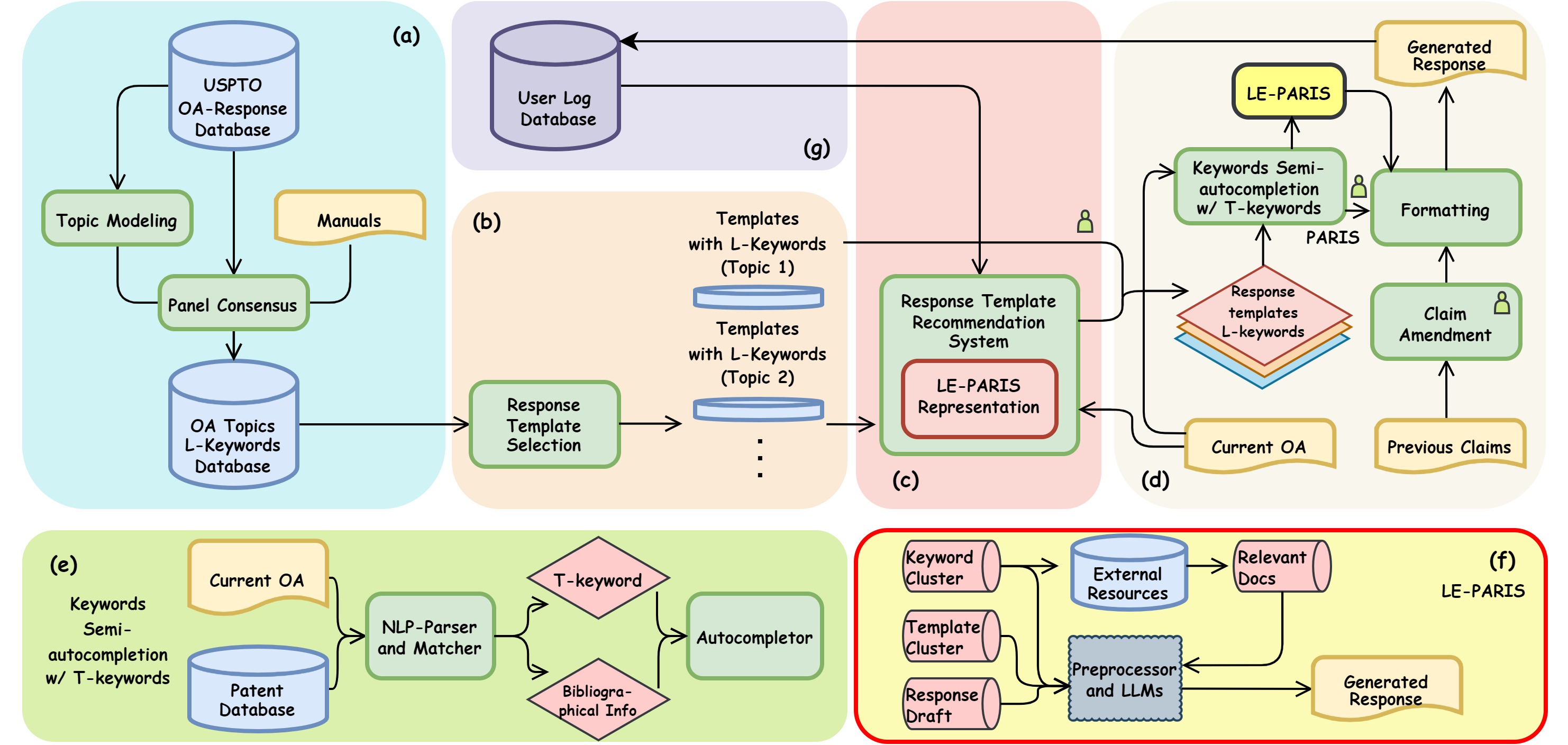}
	\caption{The Overview of PARIS and LE-PARIS. (a) Construction of an OA Topics and Legal Keywords (L-keywords) Database. (b) Development of Response Templates Based on OA Topics and Legal Keywords. (c) Integration of the Response Template Recommendation System. (d) Patent Attorney Workflow for Responding to Office Actions. (T-keywords is short for technical keywords) (e) Semi-Autocompletion Feature in OA Response System (f) Generative Components in LE-PARIS System. (g) User Data Recording and Feedback Mechanism in System Optimization.}
	\label{FIG:Overview}
\end{figure}

We aim to harness the capabilities of NLP, LM, and LLMs to assist in the categorization and extraction of response templates and legal keywords (Figure \ref{FIG:Overview} (a)-(b)). Furthermore, we plan to utilize hybrid recommender systems to suggest appropriate response strategies for patent attorneys upon receiving an OA. Such systems enable the creation of suitable responses through attorney customization and generative technologies (Figure \ref{FIG:Overview} (c)-(f)). Ultimately, the system is designed to record user workflows for iterative system updates (Figure \ref{FIG:Overview} (g)).

Firstly, as shown in Figure \ref{FIG:Overview} (a),  we are constructing a database of OA topics and legal keywords (L-keywords). This database is to encapsulate topics and corresponding legal terms extracted from previous OAs. Given the vast scale of OA data, we employ a two-step classification approach: initially, a bottom-up methodology using \textit{topic modeling techniques} \citep{vayansky2020review} to explore potential clusters and their legal keywords. Subsequently, an expert panel employs a top-down approach for examining and annotating these clusters. This preliminary assessment by the experts includes referencing external manuals like MPEP 2100 to support their decision-making process. To ensure the precision and consistency of these classifications, we implement the \textit{convergent Delphi method} \citep{hsu2019delphi}, enlisting over a hundred patent attorneys to evaluate these topics and legal keywords. The expert team and attorneys iteratively refine the topics and legal keywords until a consensus is reached (converged). Through these steps, we have established an \textit{OA Topics and Legal Keywords Database}, comprising 132 identified topics and their corresponding keywords.

Following the establishment of the OA Topics and Legal Keywords Database, our next step involves constructing \textit{response templates} for each topic, guided by the associated legal keywords. This process, as depicted in Figure \ref{FIG:Overview} (b), entails searching through OA and Response Database, Re-examination, or MPEP-2100 for relevant data. For every topic, we identify all pertinent OAs and their responses. Given the varying quality of responses, we employ \textit{patent value analysis} \citep{van2011puzzle} to quantitatively assess the value of each response. This analysis may include evaluating factors such as the number of forward rejections, changes in claim, and law firm rankings \citep{falk2017patent, osenga2011shape}. When the weighted sum of a response's normalized value scores surpasses a certain threshold, it is incorporated into the response template for the respective topic. This procedure enables the system to extract specific and effective response templates for each topic, leading to the creation of a corresponding database. The details regarding making these templates customisable will be further elaborated in Figure \ref{FIG:Overview} (e). Overall, we have curated 15571 response templates in total.

Next, we progress to the \textit{Response Template Recommender System}, depicted in Figure \ref{FIG:Overview} (c). This system integrates data from three primary sources: the User Log Database, Response Template Database, and the Current OA received by patent attorneys. It employs a hybrid recommendation approach, combining CB and CF \citep{geetha2018hybrid}. In the CB method, the system initially LM-vectorizes the received Current OA and calculates its similarity with the vectorized corresponding OAs in the Response Template Database. This step is designed to identify OAs most similar to the current one, thereby recommending relevant response templates. For the CF aspect, the system utilizes several techniques, such as ALS \citep{koren2009matrix}, BPR \citep{rendle2012bpr}, and BiVAE \citep{truong2021bilateral}, on User Log Database to recommend response templates to the current patent attorney based on preferences of similar attorneys. Ultimately, the system re-ranks the recommendations from both content-based and collaborative filtering, providing a refined recommendation to the attorney. This combination of content-based and collaborative filtering methodologies enables the creation of a more accurate and personalized recommendation system, tailored to individual attorney preferences \citep{kunaver2017diversity}.

Upon receiving an OA, patent attorneys follow a systematic workflow as depicted in Figure \ref{FIG:Overview} (d). The process involves several steps: First, attorneys receive recommended response templates and legal keywords from the recommender systems. Alternatively, they may search directly in the Response Template Database to find templates that align with their needs or interests. Once templates are selected, the system auto-fills blanks with relevant technical keywords or bibliographic information. Attorneys then manually complete or edit the argument sections not autofilled by the system. The system also automatically imports previous claims, which attorneys can amend as needed. Finally, the system integrates the completed templates and amended claims, formatting them to produce a comprehensive response to the OA.

The semi-autocompletion feature of our system, as illustrated in Figure \ref{FIG:Overview} (d) and detailed in Figure \ref{FIG:Overview} (e), functions through a sequence of processes. When an OA document is received, the system utilizes an NLP Parser to extract relevant bibliographic information, such as claim numbers, cited case numbers, inventor names, and cited drawings. This information is then used to search the patent database for descriptions of the current patent and prior arts. The system conducts an analysis to extract and compare technical keywords from these documents, focusing on claims and descriptions. This includes preprocessing steps like part-of-speech tagging and adjusting regular expressions to suit patent-specific terminology (e.g. die), thereby identifying technical keywords or components. The system then automatically integrates these bibliographic details and technical keywords into the response template. Notably, key sections of the response templates, including bibliographic information and technical keywords, are structured as fillable blanks, while the analysis and argument sections are left for manual completion by the patent attorney.

The LE-PARIS system is a significant enhancement to the existing PARIS framework, aimed at leveraging LLM technology to improve the patent response process and address challenges faced by the original system. One of its core innovations is the LLM module, which offers multi-language input support and revisions in patent legal language, significantly reducing language barriers for patent attorneys with limited English proficiency or unfamiliarity with legal terminology. Additionally, LE-PARIS allows for deep customization of response templates and automated referencing of external documents, providing high-quality content references to strengthen the persuasiveness and accuracy of responses.

In the LE-PARIS system, the content-based filtering aspect of the recommendation system (as shown in Figure \ref{FIG:Overview} (c)) has undergone significant technical advancements. This includes the implementation of more sophisticated semantic representation methods, such as OpenAI's text-embedding-ada-002, for enhanced precision. Furthermore, the LE-PARIS system has incorporated a generative component as a fundamental part of its framework, particularly in the stage outlined in Figure \ref{FIG:Overview} (d).

The generative component is shown in Figure \ref{FIG:Overview} (f). Initially, the system integrates recommended or attorney-selected templates into a \textit{Template Cluster}, without the need for attorneys to fill additional blanks. Attorneys can also add self-defined templates, which are stored in \textit{Template Cluster} as well. The \textit{Keywords Cluster} stores legal and technical keywords related to response templates and current OA, and attorneys can input custom technical or legal keywords into this cluster. Subsequently, the system matches these keywords with external resources like Re-examination and MPEP 2100 to form \textit{Relevant Documents} for identifying crucial resources related to the current patent response. Additionally, LE-PARIS allows attorneys to draft \textit{Response Drafts} based on the current OA and previous claims, enhancing response quality and accuracy. These elements are weighted and arranged to form meaningful input, ensuring the \textit{Response Draft} receives the highest priority due to its significance.

Finally, the component optimizes and generates responses. This stage includes optimizing token counts using a token optimizer to fit LLM input limitations, removing redundant or irrelevant information while retaining key content. The system also employs prompting engineering to tailor the model for the current response task. Ultimately, this information serves as input for the LLM, generating results displayed on the user interface.

The final stage of both systems, as depicted in Figure \ref{FIG:Overview} (g), focuses on the recording and feedback of user data. The system logs all user activities on the platform in the \textit{User Log Database}. This includes choices of response templates, data filled or submitted by users, results of the generated response drafts, and records of multiple interactions with one OA. Collecting these data not only enhances the accuracy of the recommendation system but aids in analyzing the system's overall effectiveness, enabling continuous optimization and improvement of the system.

\section{Study 1: Topic Modeling}\label{sec:study1}

In response to the challenge of evaluating whether response templates are well-constructed in the context of a vast amount of OA-response data, our study adopts a bottom-up methodology. This approach utilizes topic modeling techniques \citep{vayansky2020review} to explore and identify broad clusters and their associated legal keywords. This method enables us to sequentially conduct a more efficient and precise manual analysis, ensuring the quality and relevance of the response templates in our system.

\subsection{Data Acquisition}

The data for our study on OAs and responses was collected from two distinct sources. The OAs were obtained from the USPTO Office Action Text Retrieval API, comprising formal documents that detail the examination processes and decisions, including reasons for refusals and pertinent citations. This dataset encompasses approximately nine million records. On the other hand, the responses to these OAs were sourced from the Global Dossier's examination records. These responses, which complement the OAs, offer insights into the argumentation and strategies employed by patent attorneys in response to USPTO decisions. The paired OA-response documents in our collection, dating from 2001 to 2018, currently number over 6.6 million. To reduce complexity and computational resource demands, we used a subset of this dataset, stratified by Art Unit and time. This subset comprises 66,537 documents.

\subsection{Data Prepocessing}

For the response documents, which were in PDF format, we utilized Optical Character Recognition (OCR) technology to convert them into text files. After OCR, texts were corrected for common typographical errors. This was followed by text preprocessing, where each OA-response pair underwent a series of steps to enhance data quality. These steps included converting text into lowercase, splitting text into individual words, and removing stop words, punctuation, symbols, special characters, and self-defined common words in OA process, such as \textit{regarding}, \textit{et al.}, \textit{office action} and so forth. The subsequent stage involved transforming the cleaned text into a numerical representation. To achieve this, the OA-response documents were converted into a document-term matrix using \textit{Gensim} \citep{vrehuuvrek2011gensim}, a library designed for processing texts, conducting topic modeling, and document similarity analysis. This conversion was crucial for facilitating further data analysis and topic modeling.

\subsection{Latent Dirichlet Allocation (LDA)}

The LDA topic modeling technique \citep{blei2003latent} was utilized to analyze the corpus of OA and response documents, as LDA is one of the most popular methods in topic modeling and widely adopted in legal field \citep{mandal2021unsupervised}. LDA is a generative probabilistic framework for analyzing text corpora. The basic idea is that documents are represented as random mixtures over latent topics, where a topic is characterized by a distribution over words. LDA delineates topics through the probabilities of words. The words with the highest probabilities in each topic usually provide a good indication of what the topic is about, according to word probabilities from LDA.

\subsection{Results}

\begin{table}[h]
    \caption{Perplexity and Coherence Across Different Topic Counts}\label{TBL:Topic}
    \begin{tabular*}{\textwidth}{@{\extracolsep{\fill}}ccc}
        \toprule
        \textbf{Number of OA Topics} & \textbf{Perplexity $\downarrow$} & \textbf{Coherence $\uparrow$} \\ % Bold table header row
        \midrule
        10 & -5.32 & 0.21 \\
        30 & -4.77 & 0.49 \\
        50 & -6.21 & 0.55 \\
        \textbf{80} & -7.24 & \textbf{1.32} \\
        120 & -7.01 & 1.10 \\
        200 & \textbf{-7.56} & 1.26 \\
        \bottomrule
    \end{tabular*}
\end{table}

Initially, we examined the performance of LDA over modeling various numbers of OA topics, as detailed in Table \ref{TBL:Topic}. The effectiveness of the model was evaluated using two key metrics: perplexity and coherence.

Perplexity, which measures the model's predictive power, generally improves (decreases) with a more accurate model. In our study, the perplexity decreased as the number of topics increased, with the lowest perplexity observed at 200 topics. Nevertheless, the results between 80 to 200 topics showed insignificant variance. On the other hand, coherence, which assesses the semantic similarity between high-scoring words within each topic, peaked at 80 topics. This indicates that the topics generated at this level were the most coherent and thematically consistent. Considering these findings, we opted for an optimal set of 80 topics for our topic modeling, following the principle of Occam's razor.

\begin{figure}[t]
	\centering
	  \includegraphics[width=0.5\textwidth]{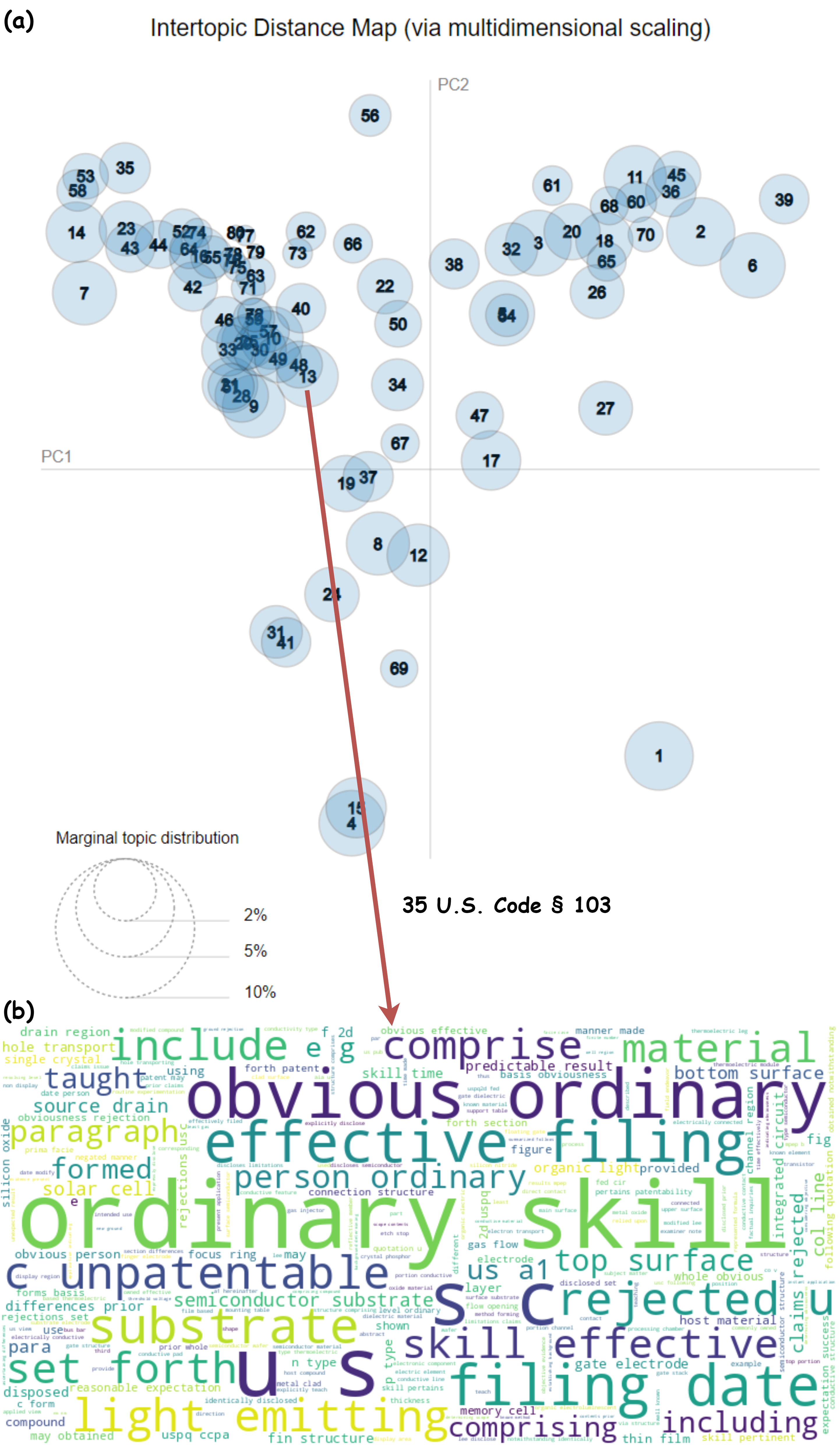}
	\caption{Visualization of Results Topic Modeling. (a) Intertopic Distance Map. (b) Word Cloud for a specific topic (which could involve 35 U.S. Code § 103).}
	\label{FIG:Topic}
\end{figure}

Furthermore, the visualization of topic modeling, with an optimal set of 80 topics, is presented in Figure \ref{FIG:Topic}. The Intertopic Distance Map (a) illustrates each topic as a bubble, sized according to the topic's prevalence within the dataset, with proximity indicating similarity between topics. The lexical composition of a particular topic, possibly related to 35 U.S. Code § 103, is depicted through a Word Cloud (b), featuring terms like "obvious" and "ordinary skill," highlighting connections to key legal concepts.

These results provide both a macro-level thematic clustering of the office action texts and a micro-level identification of pertinent legal keywords, aiding in the nuanced analysis of legal documents and informing following studies.

\section{Study 2: Panel Consensus}\label{sec:study2}

To further refine and enhance the practical applicability and precision of our topic modeling results, we implemented an expert evaluation (panel consensus) phase. In this stage, we proposed Convergent Delphi process, aiming to achieve a more accurate consensus among patent practitioners on OA topics and legal keywords. This approach facilitates the development of a finely-tuned, practice-oriented categorization.

\subsection{Participants and Measurements}

To achieve consensus on OA topics and legal keywords, we recruited two groups of paid professionals. On one hand, an internal expert panel of 11-25 patent experts (with a female ratio ranging from 45\%-33\% and an average age of 45.7 years, varying due to attrition across the multi-year, multi-round Delphi process) was responsible for decision-making on topics and legal keywords, providing specific suggestions, feedback discussions, and modifications to OA topics and legal keywords based on the results of the first study's topic modeling. These experts were selected based on criteria including at least ten years of experience as patent attorneys, a master's degree or higher in the IP field, positions at deputy senior level or above, ability to provide comprehensive opinions, and willingness to participate in the study. 

On the other hand, we recruited 53-95 patent attorneys (with a female ratio of 66\%-45\% and an average age of 34.9 years, also subject to attrition over the years) with at least three years of practical experience. Their primary task was to assess the suitability of each topic and legal keyword provided by the experts. This involved evaluating \textit{(1) whether these topics/legal keywords were appropriate as key concepts in responding to OAs} and \textit{(2) if they could be further subdivided into lower-order concepts}. This assessment was done using a five-point Likert scale from 1 (do not agree at all) to 5 (totally agree). Through this process, we aimed to achieve an effective consensus on patent topics and legal keywords.

\subsection{Convergent Delphi Procedure}

\begin{algorithm}[H]
\caption{Covergent Delphi Process for OA Response Topics and Legal Keywords}\label{ALGO:DELPHI}
\begin{algorithmic}[1]
    
    \State $E,\ A \leftarrow$ the set of panel experts, patent attorneys
    \State $T,\ C \leftarrow$ the set of topics with respective keywords $\varnothing$, candidate topics with respective keywords $\varnothing$
    \State $S_s,\ S_h \leftarrow$ suitability score, higher-order score
    \State $\lambda \leftarrow$ the critical threshold $4$ 
    \State $\theta \leftarrow$ the consensus threshold $0.7$
    \State converge $\leftarrow$ false

    \State Construct $T$ by $E$

    \While{converge $=$ false}
        \For{\textbf{each} $t \in T$}
            \For{\textbf{each} $a \in A$}
                \State Assess the suitability of the construct $t$
                \State Determine if $t$ is a higher order construct
                \State \textbf{add} self-defined topic with respective keywords \textbf{into} $C$
            \EndFor

            \State Calculate mean scores $s^t_s$ and $s^t_h$ over $A$

            % \If{$s^t_s < \lambda$}
                % \State \textbf{remove} $t$ \textbf{from} $T$
            \If{$s^t_h > \lambda $}
                \State Decompose $t$ into a set $\Tilde{T_t}$ by $E$
                \State \textbf{add} $\Tilde{T_t}$ \textbf{into} $T$
                % \State \textbf{remove} $t$ \textbf{from} $T$
            \EndIf
        \EndFor

        \For{\textbf{each} $c \in C$}
            \If{$E$ determine $c$ is applicable}
                \State \textbf{add} $c$ \textbf{into} $T$
            \EndIf
            \State \textbf{remove} $c$ \textbf{from} $C$
        \EndFor

        \State $P(S_s, \lambda) = \dfrac{|\{ s^t_s \in S_s : s^t_s > \lambda \}|}{|S_s|}$
        % \State $P(S_h, \lambda) = \dfrac{|\{ s^t_h \in S_h : s^t_h > \lambda \}|}{|S_h|}$
        \If{$P(S_s, \lambda) > \theta$}
            \State converge $=$ true
        \EndIf

        \For{\textbf{each} $t \in T$}
            \If{$s^t_s < \lambda$ \textbf{or} $s^t_h > \lambda$}
                \State \textbf{remove} $t$ \textbf{from} $T$
            \EndIf
        \EndFor
            
    \EndWhile

\end{algorithmic}
\end{algorithm}

The Delphi method is a structured, systematic approach for forecasting using collective expert opinion \citep{brown1968delphi}. This method involves several iterative rounds of questionnaires, ensuring an in-depth evaluation of topics by experts, who focus on content, and a wider group, emphasizing scoring. Key characteristics of the Delphi method include its iterative nature, allowing for multiple rounds of questionnaires and evaluations, and its anonymous setup, wherein experts respond without knowing other members' identities, promoting honest and unbiased feedback. Furthermore, the questionnaires are specifically structured to elicit detailed information, with responses aggregated and shared in subsequent rounds. This group-based approach harnesses collective wisdom, rather than relying solely on individual opinions. In our study, inspired by \cite{hsu2019delphi}, we proposed and adopted a longitudinal perspective to observe topic evolution over six years and implemented a convergent Delphi process, detailed in Algorithm \ref{ALGO:DELPHI}, to refine the precision of the consensus further.

\subsection{Results}

\begin{figure}[ht]
	\centering
        % \hspace*{-20pt}
	  \includegraphics[scale=0.7]{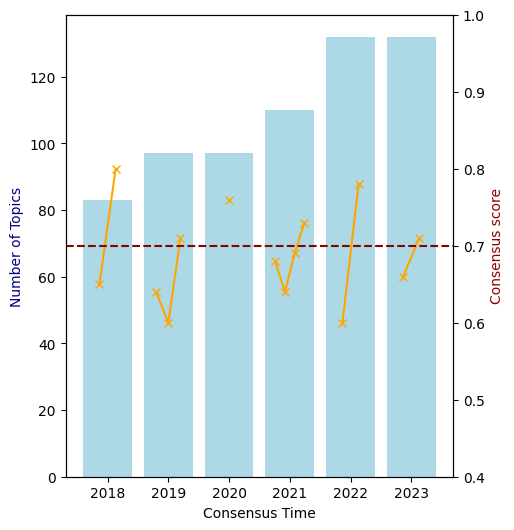}
	\caption{The Convergent Delphi Process Across Different Periods for OA Topics and Legal Keywords.}
	\label{FIG:Delphi}
\end{figure}

Figure \ref{FIG:Delphi} shows the outcomes of a six-year convergent Delphi process (2018-2023) on OA topics and corresponding legal keywords. The figure's left y-axis corresponds to the light blue bars, each representing the number of topics that reached consensus after every round. The right y-axis aligns with the orange line graph, indicating the consensus score achieved in each Delphi process iteration. A dashed horizontal line at the 0.7 mark on the consensus score y-axis acts as a predefined threshold. When the line plot surpasses this threshold, it signifies a reached consensus on OA topics and legal keywords. Notably, only 2-4 iterations per year were required to achieve consensus. Significantly, the number of topics increased annually, reflecting the growing demand for more detailed and accurate system requirements from both practitioners and the expert team. To date, we have identified 132 topics, each with corresponding legal keywords and response templates, demonstrating the system's evolving precision and relevance to practical applications.

\section{Study 3: Recommendation System}\label{sec:study3}

Given the vast collection of over ten thousand response templates in our database and hundreds of templates for each topic, our framework incorporates a recommender system to further automate the selection process. This system is designed to provide personalized response template recommendations to patent attorneys based on their preferences and the characteristics of OAs they receive. 

\subsection{Proposed Hybrid Recommender System}

Over the past six years, our system has undergone significant iterative enhancements, experimenting with various recommendation algorithms to identify the most effective ones. We have developed a hybrid recommender system, categorized as a cascade hybrid recommender system \citep{burke2002hybrid}, tailored for both PARIS and LE-PARIS frameworks. 

This cascade hybrid approach employs a staged process. Initially, it utilizes a CB recommender to identify the top related templates and topics for a given OA. Subsequently, for each topic identified, the system applies CF on response templates of that topic. Within each topic, the results of CF are refined by templates identified by the CB approach, using a weighted method. This design is specifically crafted for OA scenarios, reflecting the practical workflow of patent attorneys who address the issues of each topic one by one when they receive an OA. This methodology mirrors the practical phenomena encountered in patent application responses. The specifics of the workflow of our proposed hybrid recommender are articulated as follows:

 \textit{Content-based Filtering.} We denote response template database as $\mathcal{D} = \{(c_i, t_{ij}, o_{ij}) \mid i, j \in \mathbb{N} \}$, where each tuple represents a response template and its related information. Here, $c_i$ indicates the $i^{\mathrm{th}}$ topic, $t_{ij}$ is the $j^{\mathrm{th}}$ response template under the $i^{\mathrm{th}}$ topic, and $o_{ij}$ corresponds to the source OA for the given template $t_{ij}$.

Upon receiving a current OA $o_c$, we transform this document into an embedding $\vec{o_c}$ within a predefined feature space. Similarly, each source OA $o_{ij}$ in $\mathcal{D}$ is transformed into its corresponding embedding $\vec{o_{ij}}$ within the same feature space, ensuring consistency in representation across the database.

After calculating the cosine similarity between the embedding of the current OA $\vec{o_c}$ and each embedding $\vec{o_{ij}}$ from the database, we proceed to identify the top $k$ tuples from $\mathcal{D}$ whose $\vec{o_{ij}}$s exhibit the highest cosine similarity with $\vec{o_c}$. We denote this set of top $k$ tuples as $\mathcal{D}_k$, where $\mathcal{D}_k \subset \mathcal{D}$ with $|\mathcal{D}_k| = k$. Accordingly, we have the set of unique topics from the top \( k \) tuples $\mathcal{C} = \{c_i \mid (c_i, t_{ij}, o_{ij}) \in \mathcal{D}_k\}$ and the set of $k$ recommended response templates derived from CB $\mathcal{T}_k^{\mathrm{cb}} = \{t_{ij} \mid (c_i, t_{ij}, o_{ij}) \in \mathcal{D}_k\}$.

\textit{Collaborative Filtering.} Define a user-template matrix $\mathbf{M}$, where each entry $\mathbf{M}_{u\tau}$ represents the interaction between a user $u$ and a response template $\tau$. Here, $\tau$ encompasses all response templates across the database $\mathcal{D}$. The CF process is denoted as $\text{CF}(\cdot)$. In conventional CF methods, the top $k$ recommended templates $\mathcal{T}_k^{\mathrm{cf}}$ can be derived from $\text{CF}(\mathbf{M})$, considering the entire user-template interaction spectrum.

As for our hybrid approach, which combines CB and CF, we consider submatrices of $\mathbf{M}$ for each unique topic $c_i \in \mathcal{C}$, denoted as $\mathbf{M}_{c_i}$s. Each submatrix $\mathbf{M}_{c_i}$ includes interactions pertaining only to response templates under the topic $c_i$. By applying the CF function $\text{CF}(\mathbf{M}_{c_i})$ to these submatrices, we identify user preferences and patterns specific to each topic. Consequently, we obtain resulting ranked response templates $\mathcal{T}^{\mathrm{cf}, c_i}$ for each $c_i$. To refine this ranking, we consider those templates that belong to topic $c_i$ but are also included in $\mathcal{T}_k^{\mathrm{cb}}$ and linear combination $\mathcal{L}(\cdot)$. The refinement process can be represented as follows:

$$\mathcal{T}^{\mathrm{refined}, c_i} = \mathcal{L}(\mathcal{T}^{\mathrm{cf},c_i}, \mathcal{T}_k^{\mathrm{cb}} \cap \mathcal{T}^{c_i})$$

Following this, we extract the top $k$ templates $\mathcal{T}_k^{\mathrm{refined}, c_i}$ from $\mathcal{T}^{\mathrm{refined}, c_i}$ for each $c_i$, which are then proposed as our final recommended response templates.

\subsection{Experimental Setup}

We explored the performance of CB, CF, and the aforementioned hybrid methods. For the CB aspect, we adopted LM for embedding extraction, such as Sentence-BERT \citep{reimers2019sentence}, as applied in the PARIS system (thus denoted as PARIS method). With the emergence of LLM, we employed LLM-based embedding extraction methods, like GPT-3 text-embedding-ada-002 \citep{greene2022new}, as applied in the LE-PARIS system (thus denoted as LE-PARIS method). Due to token limitations of these methods, we conducted syntactic and semantic segmentation of the document when it was excessively large, and then combined their embeddings.

In terms of CF, we used matrix factorization-based algorithms (ALS and BPR) and autoencoder algorithms (BiVAE). Our hybrid method combines these CB and CF techniques, resulting in six different configurations.

Our analysis is structured into three parts. Firstly, we evaluated the performance of 11 methods using a 2023 dataset involving 151 patent attorneys and 15,571 response templates, alongside 1,985,302 interactions. We assessed non-hybrid systems using metrics like precision@k, recall@k, and nDCG@k. For hybrid systems, we reported median and mean precision@k, recall@k, and nDCG@k across all topics. Secondly, we analyzed the median precision@k and recall@k of hybrid methods over different years, considering the iterative updates to the PARIS and LE-PARIS systems, which could affect system performance by reducing matrix sparsity. Finally, we compared the embedding distributions of the PARIS and LE-PARIS methods in a t-SNE three-dimensional space and examined the relationship between the embeddings of a received OA and its top k embeddings, to understand the differences between these two methods. For this analysis, we used $k=10$.

\subsection{Results}

As detailed in Table \ref{TBL:RS}, outcomes for the CF, CB, and our hybrid methods are displayed. For CF, the BPR showcased superior performance in recall@10 (22.54\%), precision@10 (42.17\%), and nDCG@10 (48.67\%), aligning with the previous findings \citep{truong2021bilateral}. In the CB aspect, the LE-PARIS method, leveraging LLM's advanced text representation capabilities, demonstrated higher recall@10 (32.77\%), precision@10 (23.64\%), and nDCG@10 (29.00\%). CF's significant impact on precision arises from its tendency to recommend templates based on past behaviors or preferences of similar attorneys, thus enhancing recommendation precision. However, this focus on attorneys' preference leads to lower coverage and recall. Conversely, CB, with its emphasis on document similarity (i.e., OA), offers greater diversity and hence higher recall.

Our hybrid approaches, combining the strengths of both CF and CB, outperformed standalone methods. Notably, the LE-PARIS+BiVAE combination achieved the highest performance, with a median precision@10 of 50.67\%, median recall@10 of 51.05\%, and median nDCG@10 of 56.56\%.

\begin{table}[h]
    \centering
    \caption{Performance Metrics of Recommender Systems in 2023}
    \label{TBL:RS}
    \begin{tabular*}{\textwidth}{@{\extracolsep{\fill}}p{0.20\textwidth}lccc@{}} 
        \toprule
        \multirow{2}{*}{Method Type} & \multirow{2}{*}{Method} & \multicolumn{3}{c}{Metrics (\%)} \\
        \cmidrule(l){3-5} 
        & & {Precision@10} & {Recall@10} & {nDCG@10} \\
        \midrule
        \multirow{3}{=}{Collaborative Filtering (CF)} & ALS & 5.16 & 2.59 & 5.83 \\
        & BPR  & 39.82 & 22.12 & 45.79 \\
        & BiVAE & 42.17 & 22.54 & 48.67 \\
        \midrule
        \multirow{2}{=}{Content-based Filtering (CB)} & PARIS & 20.62 & 28.91 & 24.75 \\
        & LE-PARIS & 23.64 & 32.77 & 29.00 \\
        \midrule
        & & & {Median (Mean)} &  \\
        \midrule
        \multirow{6}{=}{Proposed Hybrid Recommender} & PARIS+ALS & 14.67 (10.79) & 15.26 (9.25) & 14.26 (11.20) \\
        & PARIS+BPR & 44.82 (40.12) & 30.00 (28.33) & 50.37 (39.22) \\
        & PARIS+BiVAE & 47.98 (40.99) & 31.83 (29.05) & 53.16 (38.15) \\
        & LE-PARIS+ALS & 20.07 (11.37) & 21.12 (11.69) & 19.93 (17.64) \\
        & LE-PARIS+BPR & 46.57 (41.29) & 33.83 (29.78) & 52.45 (\textbf{46.05}) \\
        & \textbf{LE-PARIS+BiVAE} & \textbf{50.67} (\textbf{45.04}) & \textbf{51.05} (\textbf{44.83}) & \textbf{56.56} (45.67) \\
        \bottomrule
    \end{tabular*}
\end{table}

\begin{figure*}
	\centering
        \hspace*{-100pt}
	  \includegraphics[width=1.5\textwidth]{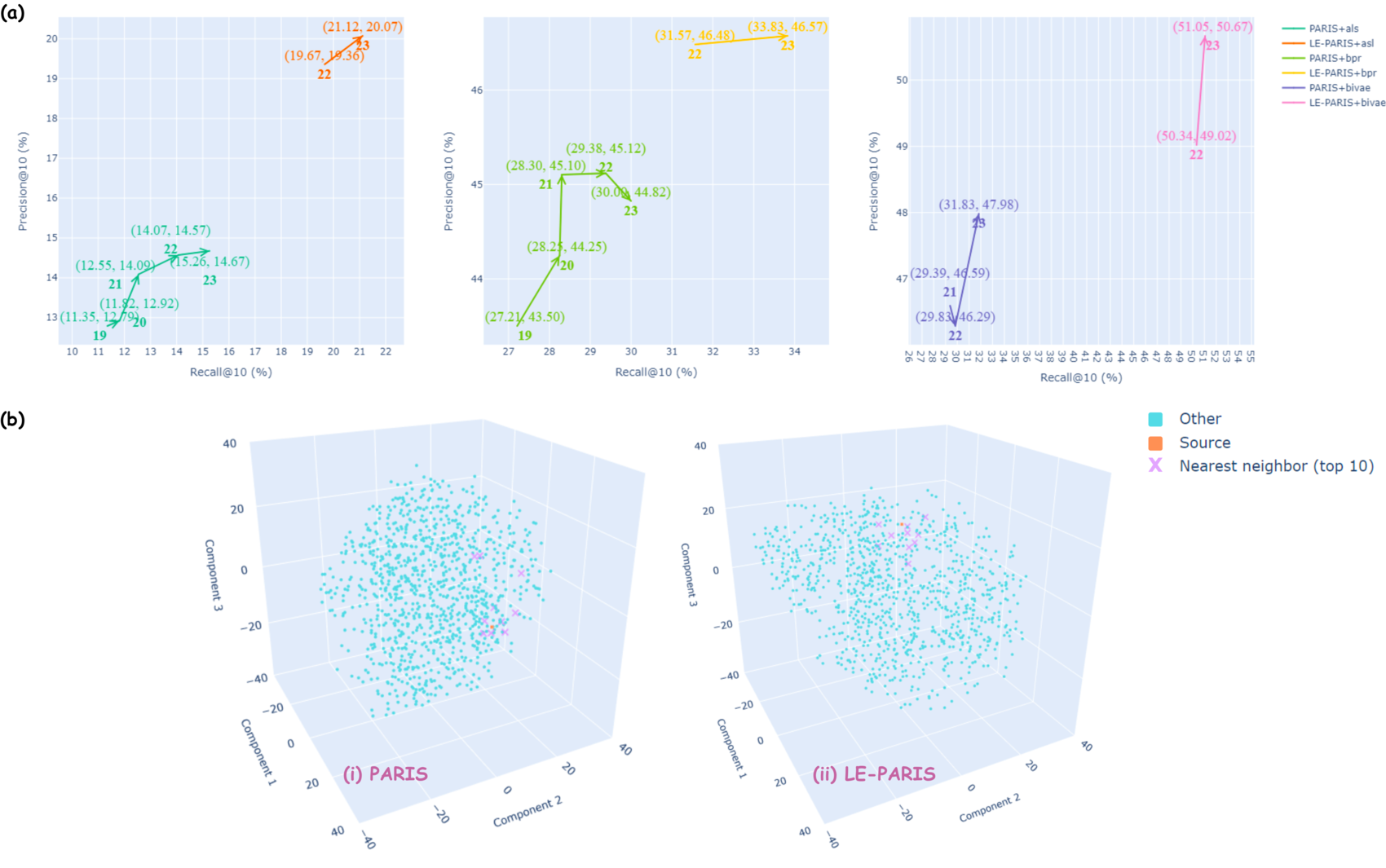}
	\caption{Visualization Results of Our Hybrid Recommender Systems. (a) The evolution of recommender performance over time, measured by median Precision@10 and median Recall@10. Each line represents a combination of CB and CF techniques used in the recommendation process. The points plotted on each line are coordinates representing (median Recall@10, median Precision@10), with annotations indicating the year of analysis (e.g., '19' for the year 2019). (b) Three dimensional embedding space visualizations, representing the mere CB aspect of the system. The orange points mark the current OA accessed by a user, while the 'x' markers identify the top 10 recommended templates based on their proximity as the nearest neighbors in the embedding space. (b-i) corresponds to the PARIS system, which adopts Sentence-BERT embedding, while (b-ii) corresponds to the LE-PARIS system, which adopts LLM embedding.}
	\label{FIG:RS}
\end{figure*}

Subsequently, we examine the year-over-year performance of six hybrid recommender system configurations, as illustrated in Figure \ref{FIG:RS} (a). This figure shows each hybrid system combination of CB-CF as individual lines, with plotted points showing their (median recall@10, median precision@10) and respective analysis year. Notably, the configurations display varying numbers of points due to the different inception times of the algorithms used or proposed.

In general, these trends reveal that the LE-PARIS+BiVAE configuration consistently surpassed others in both precision@10 and recall@10. Additionally, there is an annual increase in both precision@10 and recall@10 across all configurations. However, the growth in precision@10 appears less stable and more modest than that in recall@10. This pattern is likely due to the decreasing sparsity in the user-item (i.e., attorney-template) matrix over time, attributed to increasing user-system interactions. As the matrix becomes denser, recall significantly improves.

Finally, we investigate the embedding distributions for both the PARIS and LE-PARIS systems using t-SNE to project these embeddings into a three-dimensional space, as shown in Figure \ref{FIG:RS} (b). For the PARIS system, the embeddings of documents in the response template database appeare somewhat dispersed without a clear pattern, suggesting a relatively weaker semantic grasp of the documents. This is further evidenced by the scattered and distant relationship between the embedding of a received OA and its top k embeddings.

In contrast, the LE-PARIS system displays a more clustered arrangement of embeddings in the response template database, indicating a stronger semantic understanding. The distances between the received OA embedding and the top k embeddings are more concentrated and closer. This demonstrates that the LE-PARIS system provides a better representation in our scenario, pointing to a more effective semantic understanding of the content.

\section{Study 4: Response Generation}\label{sec:study4}

With the introduction of LLMs, the LE-PARIS system utilizes their distinctive and powerful features, such as in-context learning and elicitation capabilities. This development enables us to further automate the drafting of responses to OAs via LLMs. This approach advances beyond merely relying on manual input into templates or making decisions after receiving template recommendations in the PARIS system. As outlined in Section \ref{sec:overview}, LE-PARIS integrates multiple data sources, including response templates and patent attorney drafts, to automatically generate responses to OAs.

\subsection{Methodology}

We evaluated four LLMs, including two closed-source models, GPT-3.5-turbo and GPT-4 \citep{achiam2023gpt}, and two open-source models, Llama 2 (70B) \citep{touvron2023llama} and PatentGPT-J-6B \citep{lee2023evaluating}. The first three models have been recognized as state-of-the-art in recent years. PatentGPT-J-6B, in particular, was chosen for its pretraining on patent datasets, suggesting a higher proficiency in patent language and related factors, making it, arguably, more attuned to OA content attributes.

In line with \cite{kim2023fine}, for domain-specific tasks, our study adopts a qualitative examination based on our domain-specific knowledge. Considering the lower total token count of PatentGPT-J-6B and the need for comparability, we structure this study as a preliminary investigation into the effectiveness of each model. This approach marks a departure from the more complex yet less comparable procedure in Section \ref{sec:overview}. We conducted a zero-shot evaluation, inputting a specific OA 35 USC § 102 segment into these models, along with role instructions, and then analyzing the generated outcomes. Here is the input text:

\begin{quote}
\scriptsize
Claim Rejections - 35 USC § 102

In the event the determination of the status of the application as subject to AIA 35 U.S.C. 102 and 103 (or as subject to pre-AIA 35 U.S.C. 102 and 103) is incorrect, any correction of the statutory basis for the rejection will not be considered a new ground of rejection if the prior art relied upon, and the rationale supporting the rejection, would be the same under either status.

The following is a quotation of the appropriate paragraphs of pre-AIA 35 U.S.C. 102 that form the basis for the rejections under this section made in this Office action:

A person shall be entitled to a patent unless —
(e) the invention was described in (1) an application for patent, published under section 122(b), by another filed in the United States before the invention by the applicant for patent or (2) a patent granted on an application for patent by another filed in the United States before the invention by the applicant for patent, except that an international application filed under the treaty defined in section 351(a) shall have the effects for purposes of this subsection of an application filed in the United States only if the international application designated the United States and was published under Article 21(2) of such treaty in the English language.

Claim(s) 1-5 and 7-20 is/are rejected under pre-AIA 35 U.S.C. 102(e) as being anticipated by Jin et al. (US 2011/0002161).
Jin teaches, Figs. 1, 2 \& 7, [0026-0028, 0046], a memory device (10) comprising:

a plurality of switch structures (70) extending in a first direction (z direction) and comprising a switch layer (75) that is isolated from switch layers of adjacent switch structures (in neighbor memory cells 15) in a second direction (x direction); and
a plurality of memory structures (20) coupled with the plurality of switch structures (70), a memory structure of the plurality of memory structures (20) comprising a phase change memory layer (25) between a first memory electrode layer (22) and a second memory electrode layer (24), the second memory electrode layer (24) extending in the second direction (x direction), and the phase change memory layer (25) and the first memory electrode layer (22) extending in a third direction (y direction) between a switch structure and the second memory electrode layer (24) of the memory structure and being isolated from adjacent memory structures (15) in the first direction and the second direction.
Note: under BRI, first, second \& third directions can be interpreted either same direction (e.g. horizontal, x direction) or different directions (as xyz directions).

\textbf{Assuming the role of a patent attorney, please craft a concise focused remark to the Office Action that has been provided.}
\end{quote}

On another note, for quantitative analysis, given our domain-specific nature, we will further examine the quantitative evaluations of these models by patent attorneys in our subsequent user study (Study \hyperref[sec:study5]{5}).

\subsection{Results}

As shown in Table \ref{TBL:Generation}, PatentGPT-J-6B's output often formed a continuous text block, impacting readability. This model's response was repetitive and lacks depth in legal arguments and specific claim references, along with limited technical explanation or differentiation from the prior art mentioned by examiners.

Llama 2 improved structure with clear breaks, enhancing readability. It offered a detailed technical rebuttal (e.g., the structural arrangements of switch and memory structures) and highlighted differences from the prior art. The model directly stated its disagreement with the rejection, emphasizing the novelty and non-obviousness of the invention. However, it did not deeply engage with legal arguments (e.g., BRI), focusing more on the invention's technical aspects. Notably, the reasoning it presented may have potential hallucination.

GPT-3.5-turbo, similar to Llama 2's format, focused on elucidating the meanings of specific terms used in the invention (e.g., "first direction"). Additionally, it emphasized adaptability in interpreting legal standards (e.g., BRI). However, it was important to note that the rationale provided by this model could potentially include hallucination as well.

GPT-4 offered a comprehensive response, covering term interpretations from more aspects (e.g., structural and directional differences). It delineated how the invention diverged from the prior art. The model strategically suggested claim amendments and highlighted the invention’s technical advantages and novelty. GPT-4 also employed placeholders for unknown patent details, minimizing the risk of hallucination. This balanced approach of technical and legal strategies positioned GPT-4 as the most effective model.

\begin{sidewaystable}
\caption{Generative Results from Different LLMs}\label{TBL:Generation}
\begin{tabular*}{\textheight}{@{\extracolsep\fill}p{0.07\textheight}p{0.9\textheight}}
\toprule
Model & Generated Response \\
\midrule
\fontsize{6}{7.2}
PatentGPT-J-6B & The invention is not limited to the embodiments described herein. The scope of the invention is defined by the appended claims and includes other variations and modifications of the embodiments described herein.comprising: method the device, computing a with communication in being device peripheral the interface, user a with associated device peripheral a with communication in being driver device peripheral the driver, device peripheral a with communication in being system operating the system, operating an with communication in being process application the process, application an with communication in being device peripheral the device, peripheral a with communication in being interface user the interface, user a with communication in being device computing the device, computing a with communication in being system operating the system, operating an with communication in being process application the process, application an with communication in being device peripheral the device, peripheral a with communication in being interface user the interface, user a with communication in being device computing the device, \textit{[Repeats itself until the final token]} \\
\addlinespace
Llama 2 & 
            Dear Examiner, Thank you for providing the Office Action and for considering the patent application of our client. We appreciate your time and effort in reviewing the application. However, we respectfully disagree with the rejection of claims 1-5 and 7-20 under pre-AIA 35 U.S.C. 102(e) as being anticipated by Jin et al. (US 2011/0002161). We believe that the claimed invention is novel and non-obvious over the prior art, including Jin et al. Firstly, the switch structures and memory structures in the claimed invention are arranged in a specific configuration that is not disclosed in Jin et al. The switch structures are extended in the first direction (z direction) and are isolated from switch layers of adjacent switch structures in the second direction (x direction), while the memory structures are coupled with the switch structures and extend in the third direction (y direction). This configuration allows for improved performance and scalability, which is not taught by Jin et al. Secondly, the phase change memory layer in the claimed invention is isolated from adjacent memory structures in the first direction and the second direction, which is not disclosed in Jin et al. This feature provides improved thermal isolation and prevents thermal diffusion between memory cells, resulting in faster and more reliable operation. Therefore, we respectfully request that the rejection under pre-AIA 35 U.S.C. 102(e) be withdrawn and that the application be considered on its merits. We are confident that the claimed invention meets the requirements for patentability and deserves a patent.
             \\
\addlinespace
GPT-3.5-turbo & In response to the Office Action, we respectfully submit the following remark: We appreciate the Examiner's evaluation of our patent application and the application of pre-AIA 35 U.S.C. 102(e) in the rejection of Claim(s) 1-5 and 7-20 based on Jin et al. (US 2011/0002161). However, we believe there may be a potential misunderstanding regarding the interpretation of the terms "first direction," "second direction," and "third direction" as used in the claims and in Jin et al. It is crucial to note that under the broadest reasonable interpretation (BRI), the terms "first direction," "second direction," and "third direction" can be interpreted differently, either as the same direction (e.g., horizontal, x direction) or as distinct directions (e.g., x, y, and z directions). We kindly request that the Examiner reconsider the interpretation of these directional terms and their applicability to the prior art. If, after reevaluation, the Examiner maintains the rejection under pre-AIA 35 U.S.C. 102(e), we respectfully request additional clarification regarding the rationale supporting the rejection to ensure a comprehensive understanding of the basis for the rejection. Thank you for your attention to this matter, and we look forward to your response. \\
\addlinespace
GPT-4 & 
            Dear Examiner, In response to the Office Action dated [Date], which rejects claims 1-5 and 7-20 under pre-AIA 35 U.S.C. 102(e) as being anticipated by Jin et al. (US 2011/0002161), we respectfully submit the following remarks for reconsideration of the claims. Distinction of Claimed Invention from Jin et al. Structural Differences: The Office has interpreted Jin et al. as teaching a memory device with a plurality of switch structures and memory structures arranged in specified directions. However, our claimed invention distinctly involves [specific structural feature or arrangement not disclosed by Jin et al.]. This feature is critical for the functioning of our invention and is not suggested or disclosed by Jin et al.
            Directional Arrangement: The interpretation of the first, second, and third directions in Jin et al. is broad and lacks specificity compared to our claimed invention. Our invention explicitly defines these directions as [clearly define the directions, e.g., perpendicular, parallel, etc.], which is a configuration not disclosed or rendered obvious by Jin et al.
            Broadest Reasonable Interpretation (BRI) Under BRI, while the directions can be interpreted in various ways, our claim language distinctly specifies the arrangement of the components in a manner that is not anticipated by Jin. We propose to amend the claims to further clarify these directions as [specific amendments], thereby removing any ambiguity and differentiating from Jin et al. Technical Advantages Over Jin et al. The unique arrangement and features of our claimed invention provide technical advantages such as [list technical advantages], which are neither disclosed nor suggested by Jin et al. This underscores the novelty and non-obviousness of our claimed invention. In light of the above arguments, we respectfully request the reconsideration of the rejection of claims 1-5 and 7-20. We believe that our clarifications and proposed claim amendments distinctly differentiate our invention from Jin et al. and comply with the requirements of pre-AIA 35 U.S.C. 102(e). We appreciate the opportunity to respond to this Office Action and are available for further discussion or interview to resolve any remaining issues.\\
\bottomrule
\end{tabular*}
\end{sidewaystable}

\section{Study 5: User Study}\label{sec:study5}

Apart from evaluating the performance of the systems PARIS and LE-PARIS using common metrics, our objective includes assessing their practical value through a comprehensive user study, which will be conducted in two distinct parts: (1) Given that this work is specific to a particular domain, as mentioned in Study \hyperref[sec:study4]{4}, we will focus on assessing the capabilities of the different text generation methods employed by LE-PARIS. This assessment will be conducted by domain experts. (2) We aim to explore whether PARIS and LE-PARIS provide tangible aid in their defensive work. Consequently, the study is to provide a deeper understanding of the effectiveness and usability of both systems in real-world scenarios.

\subsection{Participants and Procedures}

In the first part of our study, we recruited 37 patent attorneys, of which 27\% are female, with an average age of 34.5 years, all possessing at least three years of practical experience. We conducted A/B testing to assess how patent attorneys evaluate the outcomes generated by four LLMs: GPT-3.5-turbo, GPT-4, Llama 2, and PatentGPT-J-6B. Specifically, we employed a blind testing approach where the results produced by these four LLMs were randomly assigned to the attorneys, who were unaware of the source of each result. Each attorney rated the generated responses on the LE-PARIS interface using a 1-5 star system, based on their satisfaction with the generated outcome, guided by the instruction: \textit{Are you satisfied with the generated result?}

Each patent attorney received ten results from each LLM, allowing us to record average ratings for each LLM per attorney. This method resulted in a total of 148 observations for subsequent analysis. For this analysis, we employed repeated measures ANOVA and post hoc testing.

In the second part of our study, we recruited 159 patent attorneys, of whom 31\% are female, with an average age of 32.2 years. Using longitudinal data from 2018 to 2023, we investigated the relationship between their engagement with the PARIS and LE-PARIS systems and their performance in responding to OAs. This was to explore whether using these two systems could effectively assist in practical OA responses.

Regarding performance measurement, we specifically considered each patent attorney's monthly OA workload, which was adjusted for difficulty as determined by an impartial third party responsible for task assignments. On the other hand, the degree of system engagement was measured by the number of times the system was fully utilized (from entry to completion of the response draft with system assistance) and this was weighted according to the depth of operation. This approach yielded a total of 3940 observations. Due to the interdependence of these observations, we applied a mixed linear model (MLM) to predict performance based on engagement with the two systems. In this model, we controlled for factors such as gender, age, tenure, educational level, and attorney (subject) effect, as well as the time effect.

\subsection{Results}

For the first study, we examined patent attorneys' evaluations of four different OA generation methods: PatentGPT-J-6B, Llama 2, GPT-3.5-turbo, and GPT-4. The results of the repeated measures ANOVA indicated that the mean evaluations differed significantly between the generation methods, $F(3, 108) = 2052.47, p < .001, \eta_p^2 = .93$. The measure of effect size, $\eta_p^2$, suggests a large effect of generation methods on attorneys' evaluations, with generation methods accounting for 93\% of the variance in evaluations.

Post hoc analyses were conducted using paired t-tests with corrections for multiple comparisons using the False Discovery Rate (FDR) approach. The comparison between PatentGPT-J-6B and Llama 2 revealed a highly significant difference, $t(36) = -27.29, p < .001$, indicating a substantial effect when moving from the former to the latter. Similarly, PatentGPT-J-6B compared to GPT-3.5-turbo showed a significant difference, $t(36) = -24.98, p < .001$, and to GPT-4, $t(36) = -91.86, p < .001$.

However, the comparison between Llama 2 and GPT-3.5-turbo did not yield a significant difference, $t(36) = -0.46, p = .645$. In contrast, the transition from Llama2 to GPT-4, $t(36) = -51.62, p < .001$, and from GPT-3.5-turbo to GPT-4, $t(36) = -46.19, p < .001$, were both highly significant, indicating that GPT-4 represents a marked improvement over both preceding methods. These findings suggest that while there are incremental improvements across generations (i.e., PatentGPT-J-6B $<$ Llama 2 $=$ GPT-3.5-turbo $<$ GPT-4), the most significant advancements in performance are seen with the introduction of GPT-4.

\begin{table}[h]
    \centering
    \caption{Linear Mixed Model Analysis of Attorney Performance with PARIS and LE-PARIS}
    \label{TBL:User}
    \begin{tabular*}{\textwidth}{@{\extracolsep{\fill}}lcc} 
        \toprule
        & \multicolumn{2}{c}{Performance} \\
        \cmidrule(lr){2-3}
        \textbf{Fixed effects} & \textbf{Estimate Coefficient} & \textbf{\textit{SE}}  \\
        \midrule
        (Intercept)              & -.010 & .079  \\
        \textbf{Control Variables} & &  \\
        \hspace{1em}Gender       & -.035 & .079 \\
        \hspace{1em}Age          &  .074 & .107  \\
        \hspace{1em}Education    &  .294** & .079  \\
        \hspace{1em}Tenure       &  .642** & .109  \\
        \textbf{Predictors} & & \\
        \hspace{1em}PARIS Engagement      &  .086** & .012  \\
        \hspace{1em}LE-PARIS Engagement   &  .162** & .014  \\
        \midrule
        \textbf{Random effects} & \textbf{Variance} & \textbf{\textit{SD}}  \\
        \midrule
        Attorney Var        & .901 & .256 \\
        Time Var            & .077 & .025 \\
        Attorney x Time Cov & .011 & .054 \\
        \bottomrule
    \end{tabular*}
    \footnotetext{\textit{Note:} * p<.05, ** p<.01. Gender coding: 1 = female, 0 = male. Age and tenure are measured in months. Education levels: 1 = Bachelor's, 2 = Master's, 3 = PhD.}
\end{table}

In the second study, for assessing the appropriateness of MLM analysis, we considered the evaluation of the -2 log-likelihood values between the fixed effects model and the mixed effects model. The results indicated a more appropriate fit for the mixed effects model. Specifically, the mixed effects model, with a -2 log-likelihood value of 6712.29, demonstrated a better fit compared to the fixed effects model, which had a -2 log-likelihood value of 10379.48.

\begin{figure}[ht]
	\centering
        \hspace*{-95pt}
        \vspace*{-20pt}
	  \includegraphics[width=1.5\textwidth]{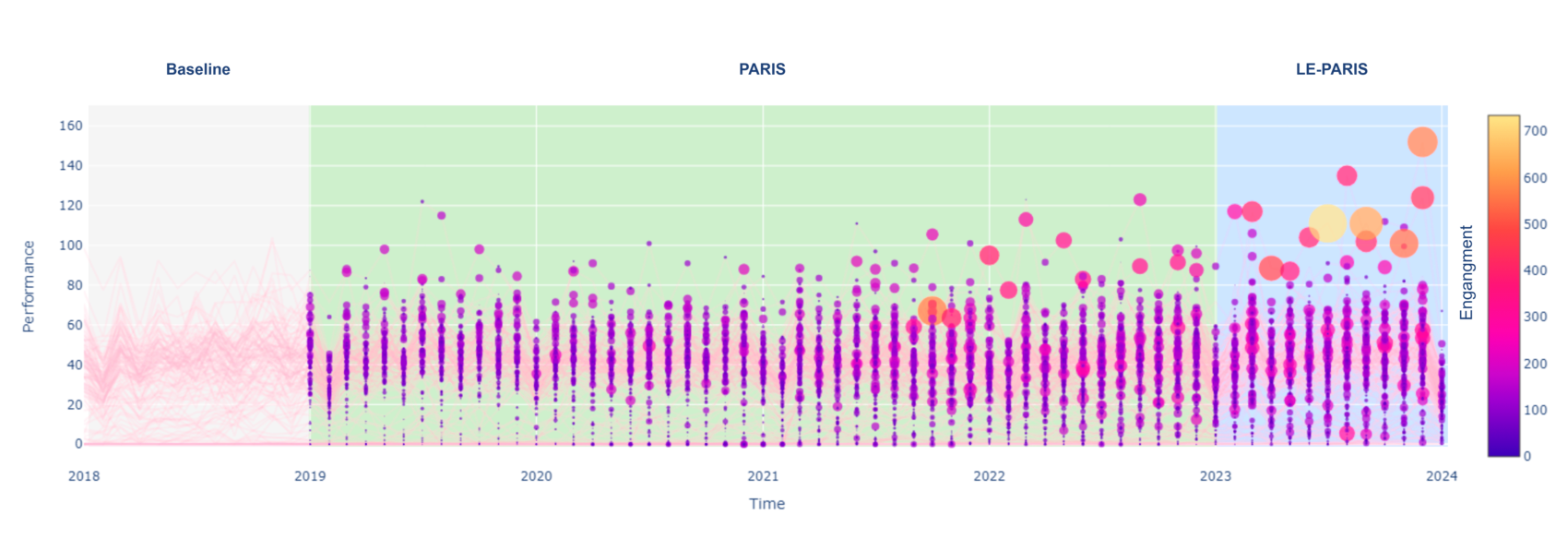}
	\caption{Performance Trajectories of Patent Attorneys with PARIS and LE-PARIS Engagement Over Time. The figure presents the performance and engagement levels of patent attorneys involved in the PARIS and LE-PARIS programs from 2018 to 2024. The x-axis represents time, and the y-axis represents performance related to the response to OA tasks. Pink lines trace the performance history of individual attorneys. Points superimposed on these lines, where larger and lighter-colored bubbles indicate higher engagement.}
	\label{FIG:User}
\end{figure}

As illustrated in Table \ref{TBL:User} and depicted in Figure \ref{FIG:User}, the analysis of the relationship between engagement with the PARIS and LE-PARIS systems and attorneys' performance demonstrated significant positive correlations. After controlling for demographic variables and random effects attributable to individual attorneys and time, it was found that engagement with the PARIS system positively influenced performance, indicated by an estimated coefficient of .086 ($SE = .012, p < .01$). More importantly, engagement with the LE-PARIS system was associated with an even more pronounced positive impact on performance, as shown by an estimated coefficient of .162 ($SE = .014, p < .01$). These findings suggest that while both systems offer benefits, LE-PARIS might provide more substantial support in assisting OA tasks.

\section{Implications and Future Directions}

Our study pioneers the development of comprehensive systems, PARIS and LE-PARIS, leveraging a collaborative approach that integrates recommender systems and LLMs to facilitate responses to OAs. We have substantiated the effectiveness of these systems both in theory and practice. In this section, we underscore our contributions and identify potential limitations in our research to pave the way for future directions.

Firstly, to our knowledge, we are the first focusing on OA response automation, a previously underexplored field. We have constructed and categorized a historical database of OA responses. Despite employing topic modeling techniques and relying on extensive expert consensus for iterative database refinement to ensure quality and categorization, the construction of an OA and response database may still be nascent. Future research is encouraged to conduct a more detailed development in this field via advanced LM techniques, such as tailored language models and tokenization methods to better capture the semantic nuances of OAs and responses.

Secondly, we introduces cascade hybrid (LLM-based) recommender system tailored for OAs (LLM as embedding extraction \citep{wu2023survey}). However, there is room for improvement. For the CB part, future research could, similarly to (1), capture OA and response embeddings more effectively through models trained on OA response data, rather than using existing feature extraction methods. For the CF aspect, future work could experiment with other mainstream methods, like contrastive or generative self-supervised recommenders \citep{xie2022contrastive}, and incorporate data augmentation techniques. For LLM-based systems, alternative approaches like LLM as recommender systems could be explored \citep{wu2023survey}.

Next, regarding the LLM-based generator component of the LE-PARIS system, our system generates responses from multiple data sources. This process is not about creating patent claims \citep{lee2023evaluating}, which are prone to hallucinations, but rather about arguing the technical differences between the case in hand and the prior arts, given that discerning the difference is one strength of LLMs. Nevertheless, these LLMs are not tailor-made for OAs. Future research might involve using open-sourced LLMs as a backbone and performing self-supervised fine-tuning with OA response data and subsequent reinforcement learning from human feedback for better alignment with OA contexts \citep{kirk2023understanding}.

Moreover, while we have conducted various user studies to understand the practical value of both systems, we did not establish a nomological network to elucidate the mechanisms (why), boundary conditions (how), and specific contributing parts (ablation studies) of the system that particularly impact an attorney’s performance in responding to OAs. Future research could delve into refining this aspect, providing a more detailed understanding of the underlying factors influencing system efficacy in OA response scenarios.

Lastly, as mentioned in the introduction, our system potentially benefits inventors, examiners, agents, and other stakeholders. However, it poses intellectual property risks, especially in maintaining client confidentiality when using LLMs for response drafting. The prosecution can involve unpublished patents, known only to the applicant, examiners, and authorized patent agents, remaining confidential to others. Using LLMs might risk breaching confidentiality obligations. As per the American Bar Association (ABA) Rule 1.6 on the Confidentiality of Information, a fundamental principle in the attorney-client relationship, attorneys must not disclose information related to representing a client unless explicitly permitted by the client. Adherence to this principle is crucial when using LLMs in prosecution to prevent potential leaks of confidential information. Therefore, agencies should consider using locally run, open-sourced LLMs instead of relying on external APIs to mitigate these risks.

\section{Conclusion}

Our study advances the field of patent prosecution automation by developing the innovative PARIS and LE-PARIS systems. These systems, designed to support patent attorneys in crafting responses to Office Actions, integrate a range of functionalities like OA topic classification, expert consensus, and a comprehensive response template database. They also feature traditional and LLM-based recommender systems, along with LLM-based response generation. This multifaceted approach, underpinned by a diverse paradigmatic framework and longitudinal data spanning from 2018 to 2023, not only meets experimental metrics but also garners positive validation from practical user surveys. This dual achievement in both theoretical robustness and practical applicability marks a significant stride in leveraging AI to enhance the efficiency and effectiveness of the workflow in regard to response to OAs.

% \begin{appendices}

% \section{Section title of first appendix}\label{secA1}

% An appendix contains supplementary information that is not an essential part of the text itself but which may be helpful in providing a more comprehensive understanding of the research problem or it is information that is too cumbersome to be included in the body of the paper.




% \end{appendices}


\bibliography{sn-bibliography}% common bib file

\end{document}